\documentclass[letterpaper, 10 pt, conference]{ieeeconf}  

\IEEEoverridecommandlockouts                              

\usepackage{graphics} 
\usepackage{epsfig} 
\usepackage{mathptmx} 
\usepackage{times} 
\usepackage{amsmath} 
\usepackage{amssymb}  
\usepackage{cite}

\usepackage[includeheadfoot,margin=2cm]{geometry}
\usepackage[font=small,labelfont=bf,tableposition=top,hypcap=false]{caption}

\def\eg{\textit{e.g.},~}
\def\ie{\textit{i.e},~}

\newcommand{\etal}{et al.~}

\title{\LARGE \bf
SplitFusion: Simultaneous Tracking and Mapping for Non-Rigid Scenes}

\author{Yang Li$^{1}$, Tianwei Zhang$^{*1}$, Yoshihiko Nakamura$^{1}$ and Tatsuya Harada$^{2,3}$
\thanks{$^{1}$ Department of Mechano-Informatics, Graduate School of Information Science and Technology, the University of Tokyo, 7-3-1 Hongo, Bunkyo-ku, Tokyo, Japan.
The first two authors contributed equally to this work. 
liyang@mi.t.u-tokyo.ac.jp; nakamura@ynl.t.u-tokyo.ac.jp}
\thanks{$^{*}$Corresponding Author: zhang@ynl.t.u-tokyo.ac.jp}
\thanks{$^{2}$ Research Center for Advanced Science and Technology, the University of Tokyo, 4-6-1 Komaba, Meguro-ku, Tokyo, Japan. harada@mi.t.u-tokyo.ac.jp}
\thanks{$^{3}$ RIKEN.}
}
\makeatletter
\let\@oldmaketitle\@maketitle
\renewcommand{\@maketitle}{\@oldmaketitle
\centering
   \includegraphics[width=1\linewidth]{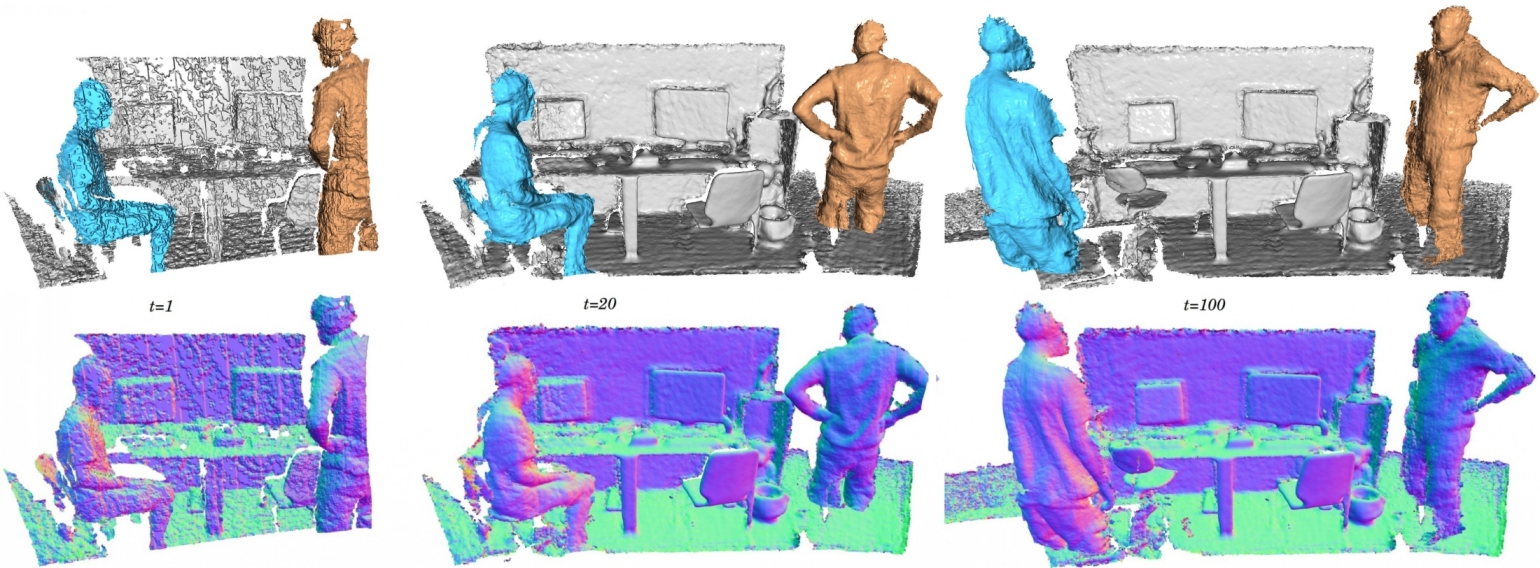}\\
 \captionof{figure}{Reconstructions of non-rigid scenes with SplitFusion; top: meshes, bottom: normal maps; both the people and the camera are moving. From left to right: frame 1, 20 and 100 of the sequence fr3/walking\_xyz from TUM RGB-D~\cite{tum_rgbd} dataset.}
\label{fig:teaser}
 }
 \let\NAT@parse\undefined
\makeatother

\usepackage[
colorlinks=true,
bookmarks=true,
citecolor=green,
linkcolor=cyan,
pagebackref=true,
pdfstartview=Fit,
breaklinks=true
]{hyperref}

\begin{document}
\maketitle
\begin{abstract}
We present SplitFusion, a novel dense RGB-D SLAM framework that simultaneously performs tracking and dense reconstruction for both rigid and non-rigid components of the scene.  SplitFusion first adopts deep learning based semantic instant segmentation technique to split the scene into rigid or non-rigid surfaces. The split surfaces are independently tracked via rigid or non-rigid ICP and reconstructed through incremental depth map fusion. Experimental results show that the proposed approach can provide not only accurate environment maps but also well-reconstructed non-rigid targets, \eg the moving humans.
\end{abstract}

\section{INTRODUCTION}
Visual Simultaneous Localization and Mapping (Visual SLAM) is a popular robotics research topic which focuses on the robot self-localization and unknown environment reconstruction using visual sensors, such as stereoscopes, monocular cameras, RGB-D sensors, and laser scanners.
Most of the existed visual SLAM approaches are designed based on the static environment assumption. 
However, dynamic elements in the scene can cause trouble for camera 6-DoF pose tracking, leading to artifacts in the reconstruction.

To deal with the problem of the dynamic environment, the recent works~\cite{Posefusion}\cite{staticfusion} extract the dynamic components from the input, removing them as exceptions to apply static SLAM frameworks. These approaches have shown improvements in both camera tracking and reconstruction. Nonetheless, the removed dynamic objects are important targets for many autonomous robot systems, such as cooperative manipulation and human-robot interactions.

The seminal DynamicFusion~\cite{dynamicfusion} proposed a general solution for non-rigid scene reconstruction by parameterizing the scene with a expandable deformation model. However, the complexity of non-rigid tracking grows quadratically with the model's size, making this approach less scalable to larger scenes. Moreover, without explicit surface segmentation, these methods can not efficiently handle topology changes.
 
In this paper, we present a novel SLAM framework called SplitFusion which simultaneously performs tracking and reconstruction for both rigid and non-rigid components of the scene.
We first split the scene into rigid or non-rigid geometric surfaces by leveraging the recent advancement in deep learning based semantic instance scene understanding.
The split surfaces are independently tracked via rigid or non-rigid ICP and reconstructed through incremental depth map fusion. The proposed visual SLAM framework not only results in accurate and clean rigid environment maps but also provides well reconstructed non-rigid objects. 

\section{Related Works}

\subsection{Online Dense 3D Reconstruction in Static Scenes}
Online dense 3D reconstruction methods require tracking and incremental integration of consecutive overlapping depth maps into a 3D representation that is continuously refined. 
Typically, camera tracking is usually performed by an Iterative Closest Point (ICP) algorithm~\cite{ICP-besl1992}. Divided by the 3D representation types, online dense 3D reconstruction has two main groups: the \textit{Surfel-Based} method and the \textit{Volumetric} method. A \textit{Surfel}~\cite{Surfels_definition} is a isolated 3D primitive with 3D coordinate, shape, and rendering attributes. For its simplicity and light computation, it has been applied in many reconstruction frameworks~\cite{PointBasedFusion}\cite{elasticfusion}. In \textit{Volumetric} method, depth maps are converted into truncated signed distance function (TSDF)~\cite{tsdf} and cumulatively averaged into a regular voxel grid. The final surface is usually extracted as the zero-value set of the implicit function using ray-casting~\cite{isosurface_raycasting}. 
KinectFusion~\cite{kinectfusion-Newcombe} is the first volumetric method that demonstrated compelling real-time reconstructions using a commodity GPU.
However, the use of a regular voxel grid imposes a large memory footprint, making KinectFusion impractical for large scale scene reconstruction. 
Nie{\ss}ner~\etal\cite{voxel_hashing} addressed the memory issue by introducing the hierarchical spatial voxel hashing method. Dai~\etal~\cite{dai2017bundlefusion} extends volumetric method into a complete SLAM system called BundleFusion, with a loop-closure, global pose optimization, and model update back-end. 

\subsection{Robust Tracking and Reconstruction in Dynamic Scenes}
The above methods assume that scenes are static. However, dynamic elements in the scene can cause trouble for camera 6-dof pose tracking, leading to artifacts in the reconstruction.
The straightforward solution is to remove the dynamic objects based on object detection and then apply the static SLAM solutions. For instance, PoseFusion uses skeleton tracking and geometry clusters to segment out humans;
DS-SLAM \cite{ds} applied SegNet to detect and remove foreground humans and then estimate the camera motion with ORB-SLAM2 \cite{orb2}.R. Martin \etal proposed Co-Fusion \cite{cofusion} and MaskFusion \cite{maskfusion} which apply object-level labels to track and reconstruct rigid objects, but they cannot handle non-rigid objects.
Some researchers insisted to find out the dynamic clusters from the dense RGB-D point clouds. StaticFusion~\cite{staticfusion} jointly refine the Visual Odometry (VO) and dynamic segmentation using intensity and depth residuals. Zhang~\etal\cite{flowfusion} proposed to involve optical flow residuals to segment non-rigid clusters from the dense point cloud clusters. Although these approaches show improvements in ego-motion estimation and background reconstruction, in many practical applications, the dynamic elements are the most important targets in the scene.  In particular, humans are widely encountered and can change the state of the scene in many ways. As such, in scenarios such as autonomous driving or the deployment of robots in areas with many humans, it would not be appropriate to discard moving elements.

\subsection{Non-Rigid Reconstruction}
The general solution for reconstruction in dynamic environments is to explicitly model the scene non-rigidity. The reconstruction of general non-rigidly deforming objects/scenes based on real-time depth sensor data has a long tradition~\cite{wand2009efficient}.
One of the traditional methods use the pre-scanned template and transforming the template to live frames from RGB-D or stereo camera data. 
Based on the volumetric approach, DynamicFusion~\cite{dynamicfusion} is the first template-free on-the-fly non-rigid 3D reconstruction system. 
Zollh{\"o}fer~\etal~\cite{volumedeform} proposed VolumeDeform, which employs a flip-flop data-parallel optimization schema that tackles the non-rigid registration at real-time rates.
Besides, they improve tracking robustness using global sparse correspondence from hand-engineered feature matching. 
The KillingFusion~\cite{killingfusion} and SobolevFusion~\cite{sobolevfusion} can handle difficult topology changes but do not recover dense spatial and temporal correspondence.   DoubleFusion~\cite{yu2018doublefusion} achieve good body tracking results by leveraging the human body model. 
SurfelWarp~\cite{surfelwarp} reduced the memory and computation cost by employs Surfel rather than a volume as the 3D representation.
Using multiple RGB-D cameras, Fusion4D~\cite{dou2016fusion4d} achieves high quality reconstruction results at the cost of the more complicated hardware setup. 
From the deep learning perspective, the recent works DeepDeform~\cite{deepdeform} replaces the noisy hand-engineered feature correspondence by more robust neural network based correspondence matching that is learned from a large amount of labeled non-rigid sequence.

In this paper, we present a novel non-rigid scene reconstruction system that by leveraging the recent advancement in semantic/instance scene understanding, to simultaneously reconstruct multiple non-rigidly deforming objects along with the static scene together.

\begin{figure*}[t]
    \centering
    \includegraphics[width=\linewidth]{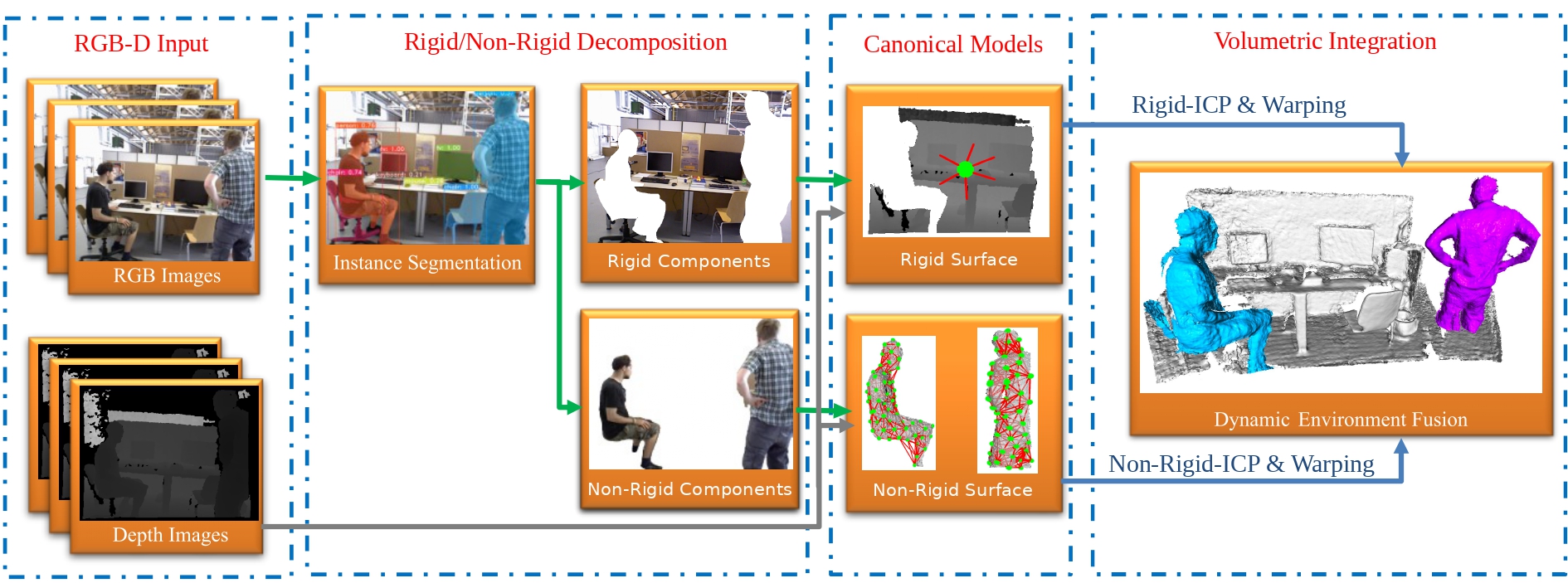}
    \captionsetup{type=figure}\addtocounter{figure}{-1}
    \caption{Overview of the proposed reconstruction system. 
For input RGB-D frames, we first apply YOLACT~\cite{yolact-iccv2019} based semantic instance segmentation on the color images and then refine the segmentation via depth map (\ie point clustering). The segmented surfaces are tracked independently.
The tracking is performed using rigid-ICP for rigid classes, (e.g. chairs and ground) or non-rigid ICP for deformable classes (e.g. human and animals).
Depth map fusion is done for each surface and the extracted meshes are combined together using the relative surface-to-camera transformations.
    } 
    \label{fig:overview}
\end{figure*}

\section{SplitFusion Overview}
SplitFusion splits a scene into several mutually independent geometric surfaces. A surface is associated with a warping filed that transforms the surface into the time-varying frames.
As show in Fig. \ref{fig:overview} the followings are the overview of this system:
\begin{enumerate}
    \item The scene is first decomposed into several independent sub-surfaces. (cf. Section.~\ref{Scene_split})
    \item The motion of a sub-surface is approximated by a dense warping filed. (cf. Section.~\ref{scene_representation})
    \item Non-linear optimization is performed to find the best warping filed parameters for each sub-surface. (cf. Section.~\ref{tracking})
    \item A live frame is fused into the warped model using volumetric method. (cf. Section.~\ref{fusion})
\end{enumerate}

\section{Technical Details}

\subsection{Scene Decomposition} \label{Scene_split}

SplitFusion takes RGB-D frames as input.
Each frame is decomposed into several pieces via deep learning based semantic/instance segmentation and the segmentation are then refined by geometric post-processing.
The system flowchart of the proposed method is shown in Fig.~\ref{fig:overview}. 
The RGB frame is first fed into YOLACT~\cite{yolact-iccv2019} to detect the object category and perform pixel-level segmentation.
YOLACT is a real-time instance segmentation method that can handle 80 types of objects in 2D images. 
Unlike two-stage methods such as Mask-RCNN, YOLACT is a one-shot instance segmentation method, which greatly reduces the inference time.  
It divide the instance segmentation into two parallel tasks. The first task uses fully convolutional networks~\cite{fcn} to generate a set of prototype masks with the same size for each image, and the output uses ReLU function for non-linearization. 
The second task is object detection based on anchor. It contains three branches: the first branch is used to predict mask coefficients for each prototype, the second branch is used to predict the confidence of instance categories and the third branch is used to predict the coordinates of the bounding box. 

There are over 80-type of objects that can be detected by YOLACT. The objects in the environments are classified into rigid or nor-rigid objects according to their semantic labels. For instance, we treat humans and animals as non-rigid,  tables, grounds, and desktops as rigid.
The next step is to split the 3D surface according to the former 2D segmentation. 

However, the YOLACT based segmentation is noisy and there are inevitable misalignments between color image and depth maps of a commodity RGB-D camera. Hence, we refine the YOLACT segmentation based on depth-map based geometry post-processing. This is done by applying the point cloud clustering as demonstrated in~\cite{Posefusion}. Specifically, the segmented non-rigid masks are first projected to 3D space using the pinhole camera model, then these projected points are used as foreground prior to start a 3D graph-cut~\cite{mincut} based point clouds splitting. 
It treats every 3D point as a vertex and the vertices are connected with their neighbors by edges. Given these mask points as foreground priors, it split the non-rigid object point clouds out from the rigid backgrounds by computing the weights of the edges.

\subsection{Per Surface Warping Filed} \label{scene_representation}
The decomposed surfaces are processed independently.
Similar to~\cite{dynamicfusion}\cite{volumedeform}, we represent the per-surface warping filed by the deformation graph $\mathcal{G}$.
In the deformation graph, the node $i$'s motion is parameterized by a translation vector $\mathrm{t}_i \in \mathbb{R}^3 $ and a rotation matrix $ \mathrm{R}_i \in \mathrm{SO}3 $. 
Therefore a node is parameterized by a rigid 6-dof transformation.
Putting all parameters into a single vector, we get the warping filed parameters
$\mathcal{G}=\{ \mathrm{t}_i  ,  \mathrm{R}_i   | i=1...k \} $, where $k$ is the total number of node.
Besides the 6-dof motion, a node also has the attribute $\mathrm{g}_i\in \mathbb{R}^3$, storing its 3D position in each time-step.
The deformation nodes are sampled in the way that it covers the surface evenly. The nodes are connected based on closeness. Refer to \cite{embededdeformation} for the details.

The motion of the entire surface is estimated by \textit{linearly blending}~\cite{embededdeformation} the motions from the graph nodes.
The influence of a rigid transformation is centered at a node's position, so that any nearby point $\mathrm{p}$ is mapped to position $\tilde{\mathrm{p}}$ according to
\begin{equation}
\label{eqn:node_centered_transformation}
\tilde{\mathrm{p} } = \mathrm{R}_i  (\mathrm{p} - \mathrm{g}_i) +\mathrm{t}_i + \mathrm{g}_i
\end{equation}
The influence of multiple graph node can be smoothly blended so that the transformation in point $\tilde{\mathrm{v}}_j$is a weighted sum of the deformation graph transformations
\begin{equation}
\label{eqn:linear_blending}
\tilde{\mathrm{v} }_j = \sum_{i=1}^{k}  w_{i,j}[\mathrm{R}_i (\mathrm{v}_j - \mathrm{g}_i) +\mathrm{t}_i + \mathrm{g}_i ] 
\end{equation}
The blending weights $w_{i,j}$ are pre-computed acoording to 
\begin{equation}
\label{eqn:blending_weights}
w_{i,j} = ( 1- || \mathrm{v}_i - \mathrm{g}_j|| / d_{max} )^2
\end{equation}
and then normalized to sum to one. Here $d_{max}$ define the max distance of a valid neighbor. The blending weights for nodes that are beyond $d_{max}$ are set to zero.
$d_{max}$ is computed per-point. It is defined by the distance to the $K+1$-th nearest neighbor node, i.e. only $K$ nearest neighbor nodes can affect a point. Empirically we use $K=6$ in our implementation.

Based on the semantic information obtained from YOLACT, we split the scene into independent rigid and non-rigid surfaces.
Note that the rigid surface can be considered as a special case of the non-rigid surface, where there are only one node in the deformation graph, \ie, it has only 6 degrees of freedom.  

In the rigid case, the non-rigid surface tracking task is reduced to the 6-dof camera pose tracking and the  $d_{max}$ is set to positive infinity, \ie all linear blending weights in Eqn.~\ref{eqn:linear_blending} become $1$.

\subsection{Warping Filed Estimation} \label{tracking}
\begin{figure}[t]
    \centering
    \includegraphics[width=\columnwidth]{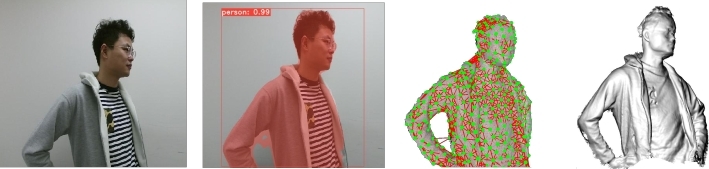}
    \caption{
    Non-Rigid surface representation. From left to right, input RGB image, semantic instant segmentation, deformation graph and surface.
    }
    \label{fig:scene_representation}
\end{figure}

\begin{figure*}[!ht]
	\centering
	\begingroup
	\setlength{\tabcolsep}{3pt} 
	\renewcommand{\arraystretch}{1} 

	\begin{tabular}{ccc}

	\includegraphics[width=.35\linewidth]{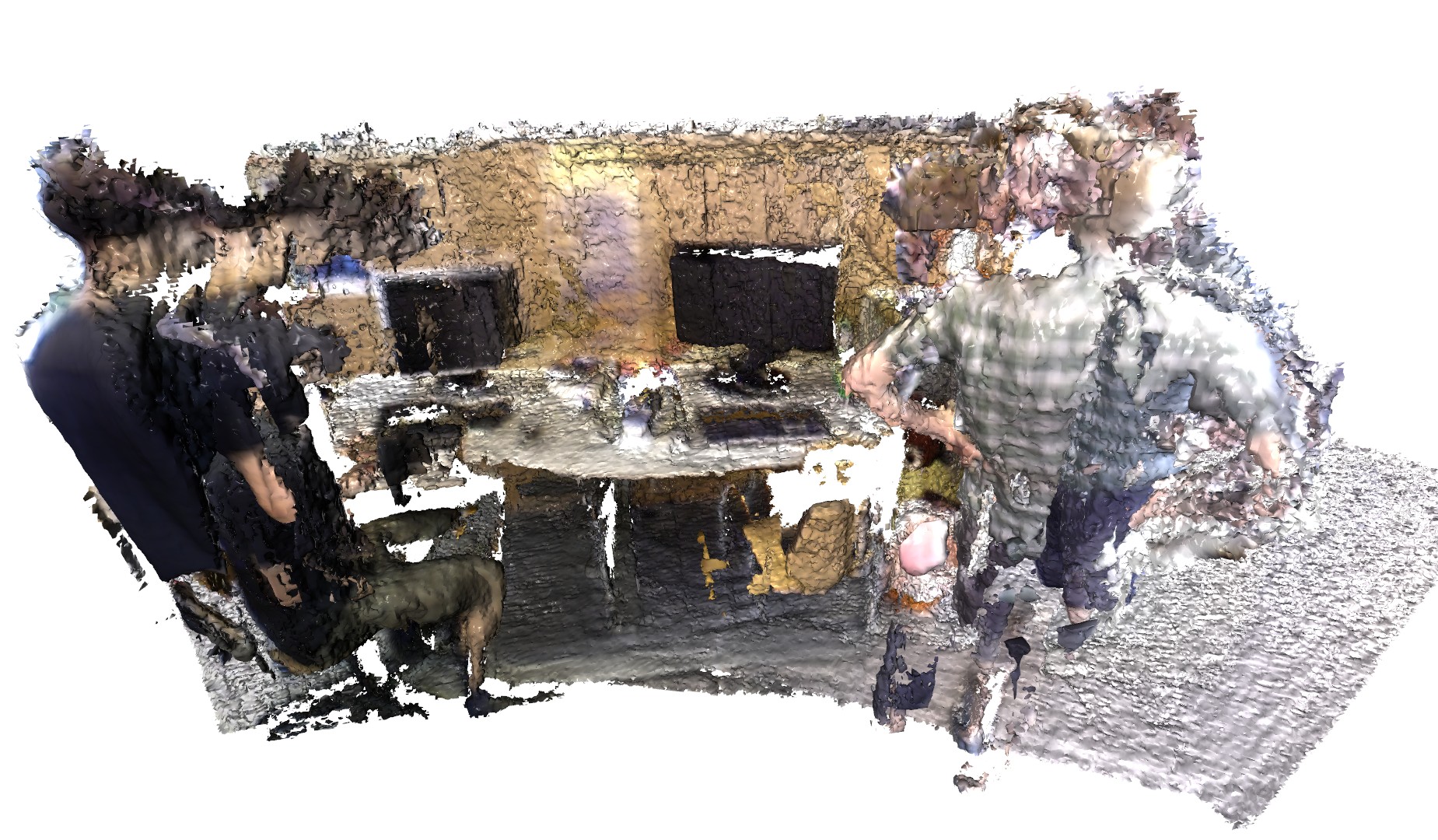} &
	\includegraphics[width=.28\linewidth]{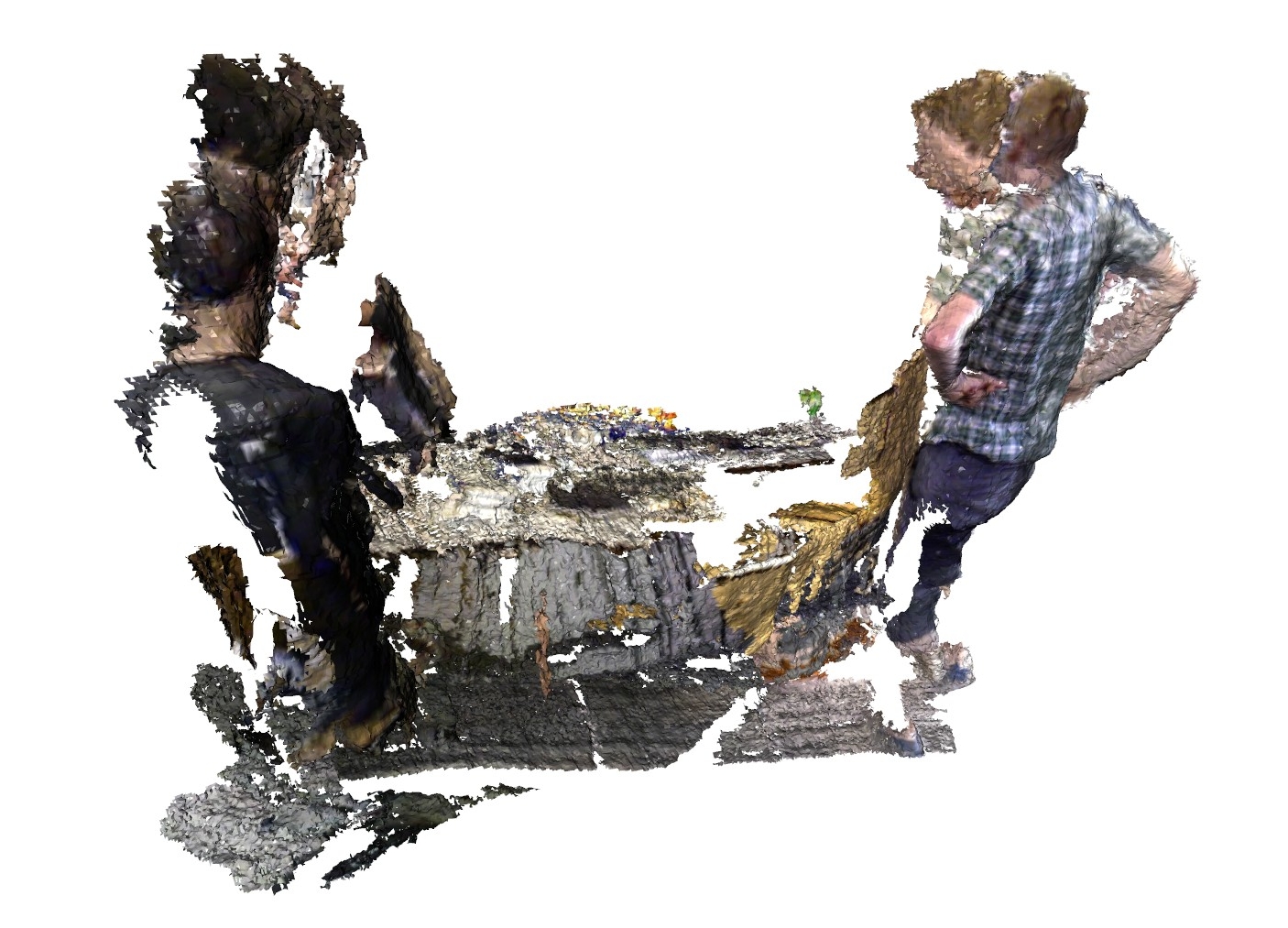} &
	\includegraphics[width=.35\linewidth]{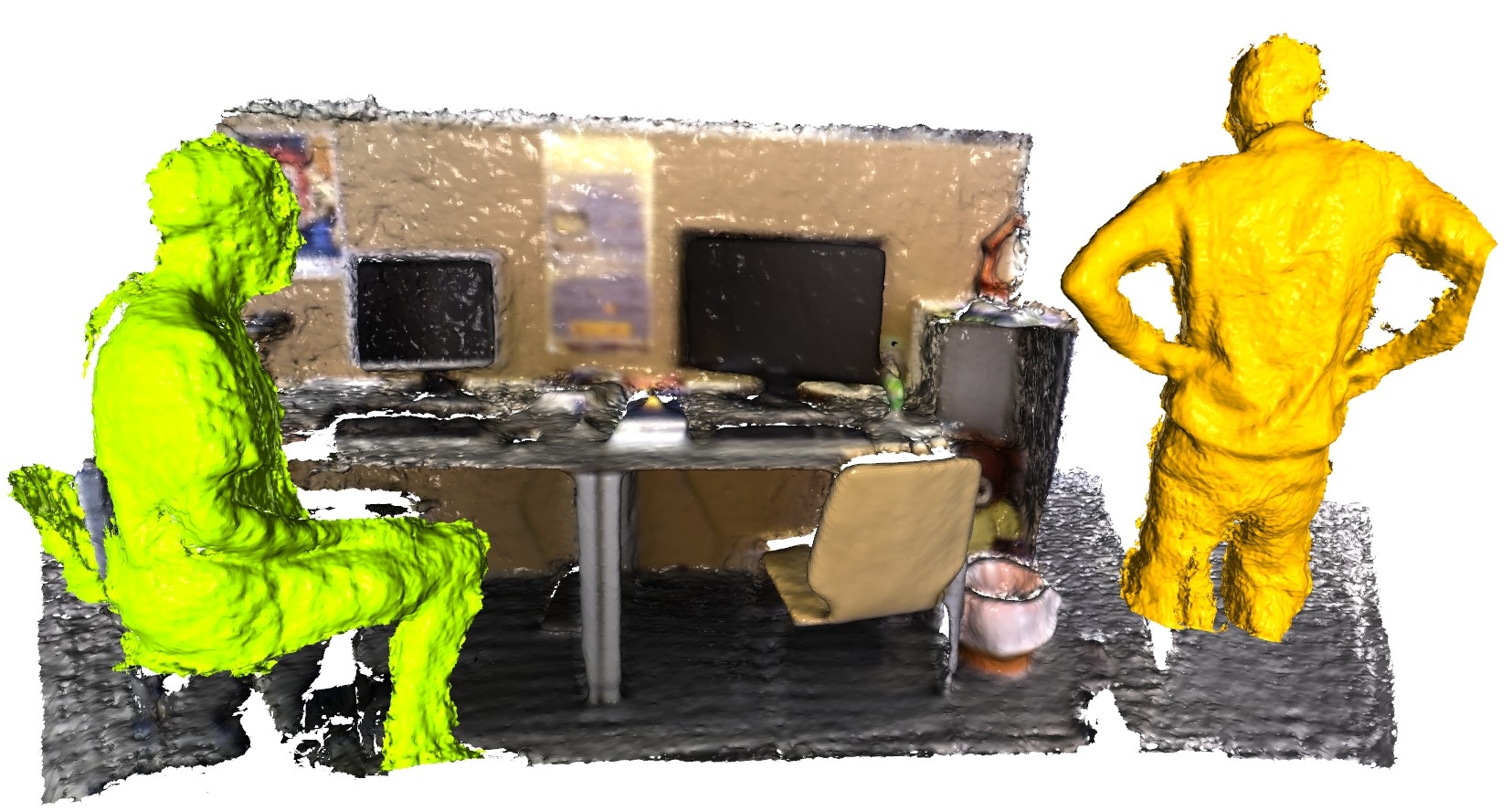} \\
	KinectFusion~\cite{kinectfusion-Newcombe} & 
    DynamicFusion~\cite{dynamicfusion} &
    Ours\\

	\end{tabular}
	\endgroup
	\caption {Reconstruction results on fr3-walking-xyz sequence from the TUM-RGBD~\cite{tum_rgbd} dataset. 
	KinectFusion does not work when the scene is non-rigid. DynamicFusion fails to deal with the topology changes without explicit scene segmentation.}
	\label{fig:tum_recon_results}
\end{figure*}

We solve the surface motion tracking problem via the non-rigid ICP approach.
For each surface, we estimate the deformation filed $\mathcal{G}$ 
given a target depth map by minimizing the following energy function:

\begin{equation}
\label{eqn:total_energy_ours}
\mathrm{E}_{total}(\mathcal{G}) = \mathrm{E}_{data}(\mathcal{G})  + \lambda\mathrm{E}_{prior}(\mathcal{G})
\end{equation}
Here the data term $\mathrm{E}_{data}$ is the dense model-to-frame ICP cost measuring the point-to-plane term between the warped model, which is a projected depth map $\mathrm{D_m}$, and live depth map $\mathrm{D_t}$. It is defined as

\begin{equation}
\label{eqn:geo_term}
\mathrm{E}_{data}(\mathcal{G}) = \sum_{(u_{m}, u_{t})\in \Omega}  
||n^T_m (v_m - v_t) ||^2
\end{equation}  
where $\Omega$ is the set of corresponding pixel pairs in the projected depth map and live depth map. $n,v\in \mathbb{R}^3$ are the normal vector and re-projected 3D vertex associated with a pixel. The method to found the correspondence set $\Omega$ can be found in the literature\cite{dynamicfusion}\cite{volumedeform}.  The point-to-plane data-term interact with the motion filed parameters according to the chain rule.


The prior term $\mathrm{E}_{prior}$ regularize the shape deformation. We use the ARAP~\cite{arap}, which encourages locally rigid motions. It is defined as
\begin{equation}
\label{eqn:prior_term}
\mathrm{E}_{prior}(\mathcal{G}) = \sum^k_{i=1} \sum_{j \in \mathcal{N}_i}  \mathcal{E}_{i,j} \cdot|| 
(\mathrm{t}_i -\mathrm{t}_j ) - \mathrm{R}_i 
(\mathrm{t}^{'}_i - \mathrm{t}^{'}_j)   
||^2
\end{equation}
Where $\mathcal{N}_i$ denotes node-$i$'s neighboring nodes, and $\mathrm{t}_j^{'},\;\mathrm{t}_j^{'}$ are the positions of $i,\; j$ after the transformation. $\mathcal{E}_{i,j}$  define the weight associated with the edge.

The energy $\mathrm{E}_{total}(\mathcal{G})$ is then optimized by the Gauss-Newton update steps:
\begin{equation}
\label{eqn:GM_step}
( \mathrm{J^TJ} ) \Delta \mathcal{G} =  \mathrm{J^Tr}
\end{equation}
where $\mathrm{r}$ is the error residue, and $\mathrm{J}$ is the Jacobian of the residue with respect to $\mathcal{G}$. 
This linear equation is solved using the data-parallel preconditioned conjugate gradient (PCG). 
Note that by splitting the whole scene into sub-surfaces, we also decomposed a potentially very large linear system into several smaller ones, which are easier to solve.
If the surface is rigid, the tracking is reduced to rigid point-to-plane ICP and the ARAP regularization disappeared from the energy function since there is only one deformation node.

\subsection{Surface Fusion} \label{fusion}
We use the TSDF function to update each sub-surface model geometry. Following the fusion technique introduced by~\cite{kinectfusion-Newcombe}\cite{dynamicfusion}, the depth maps segments $\mathrm{D_t}$ of the real-time RGB-D frame is incrementally integrated into the canonical TSDF. 
Note that each sub-surface is associated with a separate volume and the TSDF fusion is also performed independently.
We may have both rigid and non-rigid surfaces. Non-rigid surface fusion is a generalization of the projective truncated signed distance function integration approach applied in the rigid case in~\cite{kinectfusion-Newcombe}.
After the dense TSDF volumes are created, we perform per pixel ray-casting~\cite{isosurface_raycasting}  to extract the final reconstructed surfaces.  All surfaces are re-united to the camera coordinate system according to the surface-to-camera warping fileds (cf. Section~\ref{tracking}).

\begin{figure*}[t]
    \centering
    \includegraphics[width=\linewidth]{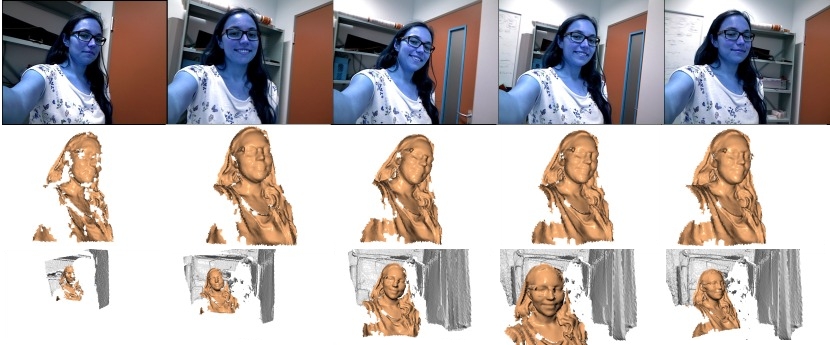}
    \caption{Our reconstruction results of the selfie sequence~\cite{staticfusion}, from left to right, frame ID: 1, 39, 72, 88, 139, 168. 
    Top row: the Color image;
    Middle row: reconstruction of the non-rigid face;
    Bottom row: reconstruction of the whole scene, including both rigid and non-rigid components.
    For the whole sequence results, we refer to the supplementary mp4 video.
    }
    \label{fig:5scona}
\end{figure*}

\begin{figure*}[t]
    \centering
    \includegraphics[width=\linewidth]{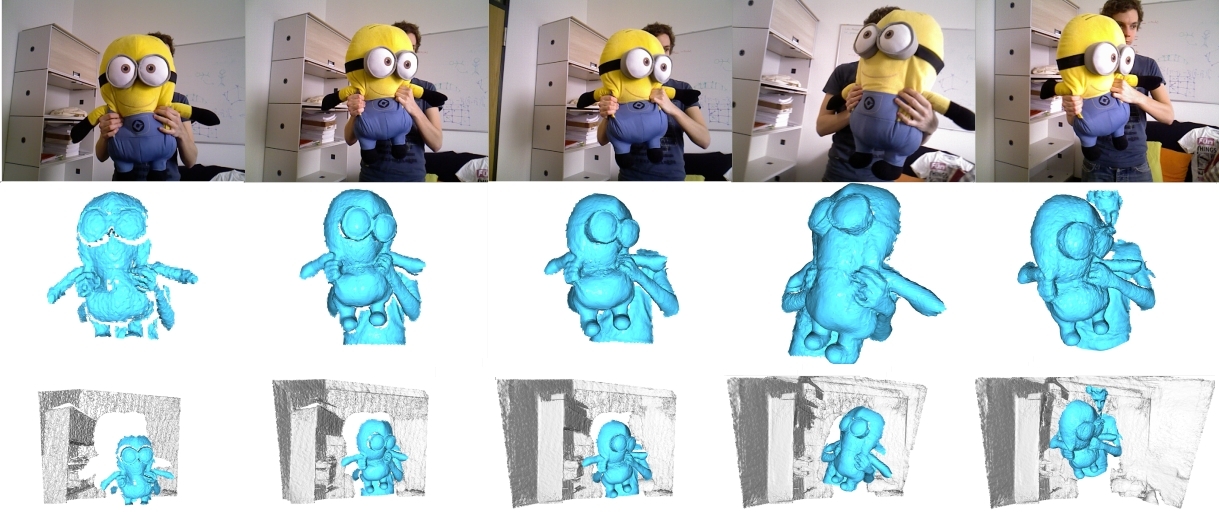}
    \caption{Our reconstruction results of the minions sequence~\cite{volumedeform}, from left to right, frame ID: 3, 50, 100, 200, 426. 
    Top row: the Color image;
    Middle row: reconstruction of the non-rigid objects;
    Bottom row: reconstruction of the whole scene, including both rigid and non-rigid components.
    For the whole sequence results, we refer to the supplementary mp4 video.
    }
    \label{fig:minions}
\end{figure*}
\section{Experimental Results}
We evaluate the proposed approach in two perspectives:
non-rigid scene reconstruction and camera pose tracking w.r.t the rigid background.
\subsection{Non-Rigid Scene Reconstruction Results}
We evaluate the reconstruction results of our method using the following three published non-rigid sequences: 
\paragraph{freiburg\_walking\_xyz} This sequence is from the TUM-RGBD~\cite{tum_rgbd} dataset. 
This scene contains two walking people in an office. It is captured via the Asus Xtion sensor which has manually been moved along three directions (xyz) while keeping the same orientation. This sequence is extremely challenging because the non-rigid human motions are very fast and the people are also interacting with the rigid objects, \ie~the chair. The sensor readings for the chairs are insufficient for detection, leading to environmental noise for tracking. 
We compare our method with the following methods: the rigid SLAM:
KinectFusion~\cite{kinectfusion-Newcombe}, the non-rigid reconstructor: DynamicFusion~\cite{dynamicfusion}, and Co-Fusion~\cite{cofusion} which also perform rigid tracking for segmented objects.
Reconstruction results for this sequence can be found in Fig.~\ref{fig:teaser} and \ref{fig:tum_recon_results}. 
KinectFusion does not work for dynamic scenes.
DynamicFusion treats the entire scene as a whole non-rigid object, the scene sticks together over time.
Although Co-Fusion also uses segmentation technique, it can not track and reconstruct the non-rigid human. We achieve significantly better reconstructions than these methods.
For video comparison, we refer to the supplementary material.
  
\paragraph{Selfie} This sequence is released by StaticFusion~\cite{staticfusion}. 
In this sequence, a person carries the camera with her right arm while the camera points at them. This a very complex test because the person often occupies more than 50\% of the image and which causes a problem for camera pose tracking and background reconstruction. The result of this sequence is shown in Fig.~\ref{fig:5scona}.
We achieve robust tracking and reconstruction for both the foreground people and the room. However, the arm and shoulder can not be reconstructed because the commodity RGB-D camera can only capture depth that is further than 0.5 meters.

\paragraph{Minion} This sequence is published by VolumeDeform~\cite{volumedeform}. This is also a very hard sequence. In this sequence, a person carries a ``Minion" in front of a camera. A large portion of the frame is occupied by the non-rigid objects. The self-occlusion between the minions and human also make non-rigid tracking very hard. For the high entanglement, we treat the person and minion as a single object.
The result of this sequence is shown in Fig.~\ref{fig:minions}. 

As shown in Fig.~\ref{fig:teaser}, Fig.~\ref{fig:5scona}, and Fig.~\ref{fig:minions} (from left to right), just like the results demonstrated in~\cite{dynamicfusion}~\cite{volumedeform}, noisy and incomplete reconstruction can be progressively denoised and completed over time as more depth maps are fused into the model. 

\subsection{Camera Pose Tracking w.r.t the Rigid Background}

To evaluate the rigid mapping results, we compared the proposed method to three state-of-the-art dynamic environment reconstruction methods: Co-Fusion(CF), StaticFusion(SF) and FlowFusion(FF). Tab.~\ref{tab1} shows the translational Absolute Trajectory Errors (ATE) on TUM\cite{tum_rgbd} and HRPSlam\cite{hrpslam} RGB-D datasets. The first two fr1 sequences are static environments, all of these methods got similar results. The lower four rows are dynamic sequences. As our method applied advanced dynamic object detection and removal technique, VO errors are much smaller than CF and competitive to SF and FF. The proposed method results in the smallest 4.8 cm ATE in fr3/walking\_xyzsequence, the reconstructed maps are shown in Fig.~\ref{fig:tum_recon_results}.
CF works well in known environments. There are moving objects in this sequence from the beginning, which result in CF's VO failure. KF and DF cannot distinguish rigid and non-rigid objects, thus they cannot decouple the camera motion and failed in non-rigid tracking.
The proposed SplitFusion split rigid and non-rigid objects. The rigid fusion pipeline provides an accurate static environment mapping and camera tracking(see the reconstructed desk and ground in the figure). Furthermore, the non-rigid fusion pipeline successfully tracks and reconstructs multiple non-rigid objects(see the yellow and green humans in the figure). 

\begin{table}[h]
\caption{Absolute Trajectory Error (ATE) RMSE (m)}
\label{tab1}  
\begin{tabular}{p{2.4cm}p{1cm}p{1cm}p{1cm}p{1cm}}
\hline\noalign{\smallskip}
\textbf{Sequence}  & CF & SF & FF & Ours  \\
\noalign{\smallskip}\hline\noalign{\smallskip}
      fr1/xyz & 0.014& 0.017 &0.020 &0.015\\
      fr1/desk2 & 0.17& 0.051 & 0.034& 0.040 \\
\noalign{\smallskip}\hline\noalign{\smallskip}
fr3/walk\_xyz& 0.71  & 0.21 & 0.12 & 0.048\\
fr3/walk\_static & 0.58 & 0.031 &0.23 &0.21\\
HRPSlam2.1 & 0.91 & 0.25 &0.23 &0.22\\
HRPSlam2.4 & 0.63 & 0.44 &0.49 & 0.45\\ 
\noalign{\smallskip}\hline\noalign{\smallskip}
\end{tabular}
\vspace{-0.5cm}
\end{table}

\section{Conclusions and Limitation}
In this paper, we proposed SplitFusion, which simultaneously tracks and reconstructs the rigid backgrounds and the deformable moving objects. 
By leveraging the learning-based semantic/instance scene understanding, our camera 6-dof pose tracking performance is comparable to state-of-the-art dynamic SLAM solutions. Furthermore, the proposed approach can track and reconstruct multiple non-rigid object surfaces and mapping them to the same world coordinate system. Experimental results show that we can reconstruct accurate static environment maps and multiple non-rigid objects even in the challenging dynamic scenes. 
However, many problems are yet to be solved: 1)the system requires excessive computation and memory on complex scenes with multiple objects, this problem could be alleviate by using light surfel as 3D representation. 2) Both rigid and non-rigid tracking are not reliable when facing large motions, a promising direction is to incorporate the robust deep trackers as show in~\cite{li2019pose}\cite{deeptam}\cite{deepdeform}\cite{yang2020}\cite{bovzivc2020neural}.

{\small
	\bibliographystyle{IEEEtran}
	\bibliography{egbib}
}
\end{document}